%% file: main.tex
\documentclass[9pt,conference]{IEEEtran} 
\IEEEoverridecommandlockouts 
\input{setting-ieee}

\graphicspath{{./figures/}}

\usepackage[T1]{fontenc}
\usepackage{xurl}
\usepackage{ragged2e}
\usepackage{microtype}
\usetikzlibrary{matrix}

\begin{document}

\title{Meta-RTL: Evolving Agent Harnesses for RTL Optimization}

\author{
    \IEEEauthorblockN{
        Xinzhuang Xiong, Yonghao Wang, Shengyao Sun,
        Chengxi Li, and Evangeline Young
    }
    \IEEEauthorblockA{
                Department of Computer Science and Engineering,
        The Chinese University of Hong Kong
    }
}

\maketitle
\pagestyle{plain}
\thispagestyle{plain}

\begin{abstract}
Large language models enable agents to optimize RTL using EDA feedback, yet their executable search procedures remain largely hand-engineered. We present Meta-RTL, an EDA-grounded framework that evolves RTL optimization harnesses from a minimal loop while keeping the optimization model, EDA flow, and per-design budgets fixed. Compared with Dr.~RTL, Meta-RTL$^\ast$ reduces aggregate WNS/TNS violations by 7.8\%/11.0\% across 20 development designs, with area reductions on 70\% of them. Without further evolution, Meta-RTL$^\ast$ transfers to 10 unseen IC-RTL designs, reducing aggregate WNS/TNS violations by 20.5\%/14.0\% and total synthesized area by 12.6\%. We further characterize three search mechanisms in the evolved harness and assess their contributions through ablations and a case study. 

\end{abstract}

\input{sections/introduction}
\input{sections/related_work}

\input{sections/method}

\input{sections/experiments}

\input{sections/limitations}
\input{sections/conclusion}

\section*{Content Generated by AI}

ChatGPT (OpenAI) was used to assist with drafting
and refining the paper,
as well as generating and revising portions of the
released codebase. All AI-assisted content was
reviewed by the authors, who take
full responsibility for the final work.

\clearpage 
\begingroup
\RaggedRight
\bibliographystyle{IEEEtran}
\bibliography{references}
\endgroup

\end{document}

%% file: setting-ieee.tex
\usepackage{blkarray}                                      
\usepackage{graphicx}                                      
\usepackage{amsmath}
\usepackage{amssymb}
\usepackage{amsfonts}
\usepackage{amsthm}
\usepackage[mathcal]{eucal}
\usepackage{mathrsfs}
\usepackage{booktabs}
\usepackage{enumerate}
\usepackage{multirow}
\usepackage[listofformat=subsimple,subrefformat=simple,farskip=0pt,justification=centering]{subfig}
\usepackage{color}
\usepackage{cite}                                          
\usepackage{comment}                                       
\usepackage{soul}                                          
\soulregister\cite7
\soulregister\ref7
\soulregister\pageref7
\usepackage{etoolbox}                                      
\usepackage{url}
\usepackage{nth}                                           
\usepackage{bm}                                            
\usepackage{courier}
\usepackage{balance}
\usepackage{threeparttable}
\usepackage{xcolor,colortbl}
\usepackage{footnote}
\usepackage{listings}
\usepackage{setspace}                                      
\usepackage[inline]{enumitem}
\usepackage{verbatim}
\usepackage[bookmarks=false]{hyperref}
\hypersetup{
    colorlinks = true,
    citecolor  = blue,
    linkcolor  = blue,
    urlcolor   = blue,
}
\usepackage{tikz}
\usetikzlibrary{patterns,snakes}
\usetikzlibrary{positioning,calc,fit,decorations.pathmorphing,shapes.geometric, shapes.gates.logic.US, calc}
\usetikzlibrary{arrows,arrows.meta,decorations.markings,shapes,shapes.arrows}
\usetikzlibrary{decorations,decorations.pathreplacing}
\usetikzlibrary{backgrounds}
\usepackage{filecontents}                                  
\usepackage{pgfplots}
\usepackage{pgfplotstable}
\usepgfplotslibrary{groupplots}
\usepackage{scalefnt}
\pgfplotsset{compat=newest}
\usepackage{caption}
\usepackage{pifont}                                        
\usepackage{cleveref}
\Crefformat{figure}{Fig.~#2#1#3}                           
\crefformat{figure}{Fig.~#2#1#3}
\Crefname{subfigure}{Fig.}{Figs.}
\crefname{figure}{Fig.}{Figs.}
\Crefname{figure}{Fig.}{Figs.}
\Crefname{equation}{Eq.}{Eqs.}
\crefname{equation}{Eq.}{Eqs.}
\crefname{table}{Table}{Tables}
\Crefname{table}{Table}{Tables}
\crefname{section}{Section}{Sections}
\Crefname{section}{Section}{Sections}
\crefname{subsection}{Section}{Sections}
\Crefname{subsection}{Section}{Sections}
\crefformat{table}{Table~#2#1#3}
\Crefformat{table}{Table~#2#1#3}
\usepackage[figuresright]{rotating}

\usepackage{algorithm}
\iftrue
\usepackage{algpseudocode}                                 
\algrenewcommand\textproc{\texttt}
\makeatletter
\let\OldStatex\Statex
\renewcommand{\Statex}[1][3]{%
  \setlength\@tempdima{\algorithmicindent}%
  \OldStatex\hskip\dimexpr#1\@tempdima\relax
}
\makeatother
\else
\usepackage{algorithmic}
\fi

\definecolor{CUHKorange}{RGB}{244,106,18} 
\definecolor{CUHKblue}{RGB}{0,111,190}    
\definecolor{CUHKgreen}{RGB}{0,127,128}   
\definecolor{CUHKred}{RGB}{228,46,36}     
\definecolor{CUHKyellow}{RGB}{198,148,34} 
\definecolor{CUHKdark}{RGB}{114,44,114}   
\definecolor{CUHKmiddle}{RGB}{144,44,144} 
\definecolor{CUHKlight}{RGB}{167,44,167} 
\definecolor{CUHKpurple}{RGB}{117,15,109}
\definecolor{CUHKgold}{RGB}{221,163,0}
\definecolor{CUHKribbon}{RGB}{244,223,176}
\definecolor{CUHKblack}{RGB}{34,24,21}

\renewcommand{\bf}[1]{\textbf{#1}}
\renewcommand{\tt}[1]{\texttt{#1}}

\usepackage{tcolorbox}
\tcbuselibrary{skins,breakable}
    {\endtcolorbox}
    {\endtcolorbox}

\usepackage[top=0.75in,bottom=1in,left=0.625in,right=0.625in]{geometry}   
\crefname{mytheorem}{Theorem}{Theorems}
\crefname{mylemma}{Lemma}{Lemmas}
\crefname{myclaim}{Claim}{Claims}
\crefname{myproperty}{Property}{Properties}
\crefname{mycorollary}{Corollary}{Corollaries}

\RequirePackage[normalem]{ulem} 
\RequirePackage{color}\definecolor{RED}{rgb}{1,0,0}\definecolor{BLUE}{rgb}{0,0,1} 

%% file: sections/introduction.tex
\section{Introduction}
\label{sec:introduction}

Optimizing register-transfer-level (RTL) designs for
timing and area is a critical task in digital hardware
design. Functionally equivalent RTL descriptions can
synthesize into substantially different circuits, so the
way a design is written directly determines its timing
and area. Synthesis tools~\cite{yosys,abc} apply predefined
optimization passes and heuristics, but they optimize
within the structure that the RTL expresses rather than
changing it. Designers therefore restructure RTL manually,
with each revision followed by functional verification,
synthesis, and timing--area analysis. This iterative
process is labor-intensive and difficult to scale as
designs grow more complex.

Large language models (LLMs) have opened new
opportunities to automate this process. Early studies
generated RTL from high-level specifications, improving
generation quality through domain-specific fine-tuning
and functional validation~\cite{verigen,rtlcoder,vericoder}.
Later approaches fed compilation, simulation, and PPA
results back to the model to iteratively refine generated
designs~\cite{autochip,llm_verippa}. Recent agentic
frameworks go further, coupling LLM reasoning with EDA
tools to generate, verify, and optimize RTL designs
autonomously~\cite{verilogcoder,veriopt,symrtlo,evolve,dr_rtl,rtlscout}.

For such agents, how the search is organized matters as
much as which transformations the LLM proposes. Every
candidate must remain functionally equivalent, timing and
area often conflict, and each evaluation requires synthesis
and equivalence checking. The search is governed by the
agent's \emph{harness}: the executable procedure that
presents EDA feedback to the model, decides which
candidates to keep, and allocates the evaluation budget.
It therefore shapes both optimization quality and
cost~\cite{ares}. Existing agents differ precisely in these
choices, from predefined workflows for tool invocation and
iterative rewriting~\cite{veriopt,symrtlo,evolve} to
persistent memory, reusable skills, and shared candidate
pools~\cite{dr_rtl,rtlscout,veriagent}. Yet all of these
harnesses are designed by hand, and each embeds fixed
assumptions about how optimization should proceed.
Promoting only the best-scoring candidate, for example,
assumes that good designs are reached through steady
improvement, whereas profitable restructuring often passes
through intermediate designs that look worse before they
are refined. In these agents, EDA feedback improves the
designs but not the harness that produces them.

\begin{figure}[!t]
    \centering
    \setlength{\abovecaptionskip}{3pt}
    \includegraphics[width=\columnwidth]{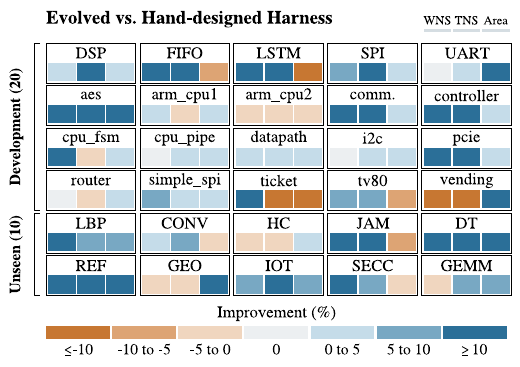}
    \caption{
Meta-RTL$^\ast$ versus Dr.~RTL on 20 development
and 10 unseen designs. Triplets show relative reductions
in WNS/TNS violations and area;
blue/orange indicates improvement/regression.
    }
    \label{fig:harness-teaser}
\end{figure}

This raises a natural question: Without manually prescribing sophisticated search heuristics, can we evolve RTL optimization harnesses from EDA feedback that outperform hand-designed ones—and if so, what makes the evolved harnesses effective?

To answer these questions, we present \textbf{Meta-RTL},
which evolves an RTL optimization harness from a minimal
seed through two nested loops. In the inner loop, a fixed
LLM optimizes RTL designs under the current harness. In the outer loop, a proposer LLM improves the harness much
as an engineer would: it traces EDA failures in the inner-loop
logs back to specific harness decisions, hypothesizes a
targeted fix and its expected effect, probes the fix on a few
relevant designs, and revises it based on the evidence. Each
submitted harness is then frozen and scored on all development
designs by synthesis and equivalence checking outside its
control. In this way, EDA feedback improves not only the designs but
also the harness that produces them. While skill libraries
capture \emph{what} transformations to apply, Meta-RTL
learns \emph{how} to search for effective transformations,
turning a traditionally hand-designed process into
an optimization target.

\begin{figure*}[t]
  \centering
  \setlength{\abovecaptionskip}{3pt}
  \includegraphics[width=\textwidth]{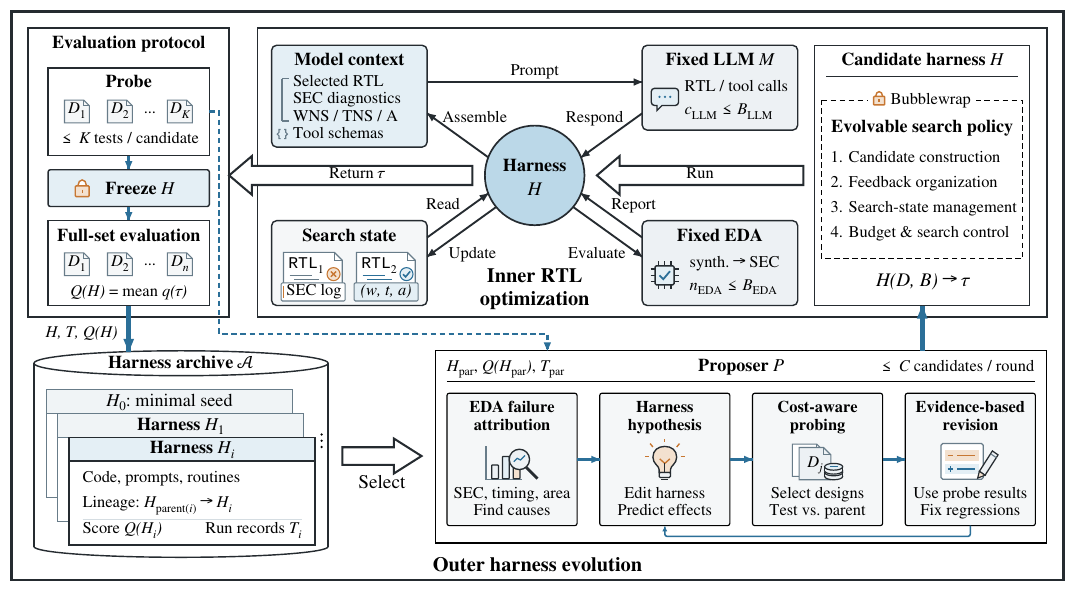}
  \caption{Overview of Meta-RTL: inner-loop RTL optimization and outer-loop harness evolution. }
  \label{fig:meta_rtl_overview}
\end{figure*}

The main contributions of this work are summarized as follows:
\begin{itemize}
    \item We formulate EDA-grounded harness evolution for
    RTL timing--area optimization, in which the search
    procedure itself evolves from a minimal loop under a
    fixed LLM, EDA flow, and budget per design. Its proposer
    revises harnesses through four designed stages described
    in Section~\ref{sec:method_evolution}.

    \item We show that harness evolution yields substantial
    timing and area gains (Fig.~\ref{fig:harness-teaser}).
    Compared with Dr.~RTL, Meta-RTL$^\ast$ reduces aggregate
    WNS/TNS violation magnitudes by 7.8\%/11.0\% and area
    on 14 of 20 development designs (Section~\ref{sec:exp_main}). Without further
    evolution, it also improves WNS, TNS, and area on 7, 8,
    and 7 of the 10 unseen designs, respectively (Section~\ref{sec:exp_generalization}).

    \item We identify three search mechanisms that emerge in
    the evolved harness, and further assess their contributions
through targeted ablations and a circuit-level case study (Sections~\ref{sec:ablations} and~\ref{sec:case_study}).
\end{itemize}

%% file: sections/related_work.tex
\section{Related Work}
\label{sec:related_work}

\subsection{Conventional Synthesis and RTL Optimization}

Traditional hardware optimization relies on
logic synthesis and physical design tools to
improve area and timing. Early synthesis systems
such as SIS~\cite{sis} introduced systematic
logic and sequential circuit optimization.
Yosys~\cite{yosys} provides an extensible
framework for RTL synthesis, while
ABC~\cite{abc} supports AIG-based logic
optimization and technology mapping.
OpenROAD~\cite{openroad} enables automated
physical design through placement, routing,
and timing optimization. These tools employ
well-established optimization algorithms, but
high-level RTL restructuring still relies on
manual design expertise.

\subsection{LLM-Based RTL Optimization}

Recent studies leverage LLMs for iterative RTL
optimization, in which candidate designs are
repeatedly rewritten and evaluated with EDA
feedback~\cite{symrtlo,evolve,poet}.
Dr.~RTL~\cite{dr_rtl} introduces reusable
optimization skills distilled from tool feedback.
RTLScout~\cite{rtlscout} instead exploits shared
candidate pools and experience accumulated across
runs. Despite these advances, the executable
procedures organizing optimization remain largely
hand-engineered.

Building on broader progress in self-evolving
systems~\cite{harness_evolve,meta_harness,self_harness},
recent studies have explored program evolution for
EDA tasks. These include logic
synthesis~\cite{evoabc}, technology
mapping~\cite{mappingevolve}, and physical design
tasks such as placement~\cite{goal_evolve,coevopr},
routing~\cite{gr_evolve}, and FPGA
place-and-route~\cite{vpr_evolve}. Similar approaches
have also been applied to hardware
design~\cite{horizon} and
verification~\cite{hardware_harness_evolution}.

Unlike approaches that evolve internal EDA heuristics, Meta-RTL evolves the executable search procedure of RTL optimization agents. We investigate whether synthesis and equivalence-checking feedback can guide this evolution to outperform hand-engineered agents and generalize to unseen designs.

%% file: sections/method.tex
\section{Meta-RTL}
\label{sec:method}

\subsection{Overview}
\label{sec:overview}

As shown in Fig.~\ref{fig:meta_rtl_overview},
Meta-RTL evolves executable RTL optimization harnesses
from a minimal loop $H_0$ that simply edits RTL and evaluates
the resulting
designs using default EDA reports. An agentic proposer $P$
progressively revises the harness using the
implementations and execution evidence of earlier rounds, and
 the evaluation protocol specifies how it interacts with the fixed evaluator.

For an RTL design $D$ and resource budget $B$, a harness
$H(D,B)$ defines an inner optimization procedure that coordinates
a fixed LLM $M$ with equivalence checking, synthesis, and timing
analysis. Its execution produces a trajectory $\tau$ containing
RTL candidates, tool outcomes, timing--area measurements, and
search decisions. The harness determines how candidates are
constructed, how feedback is presented, and how subsequent
transformations are explored.

An archive $\mathcal{A}$ stores harness implementations,
evaluation results, and execution trajectories.
Harnesses are ranked by normalized timing--area utility
$Q(H)$ (Section~\ref{sec:method_evaluation}), and the highest-scoring eligible
harness, Meta-RTL$^\ast$, is selected and then evaluated on unseen designs.

\subsection{Harness Representation and Search Space}
\label{sec:harness_space}

A harness $H$ is a self-contained Python package that orchestrates the inner optimization loop via two primitives: invoking the fixed LLM $M$ and querying the fixed EDA evaluator for synthesis, timing, and sequential equivalence checking (SEC). $H$ can freely define prompt templates, helper scripts, and search routines, through which evolution may modify four aspects of the search procedure: (1)~candidate construction, (2)~feedback organization, (3)~search-state management, and (4)~resource allocation and search control. 

Harnesses execute in Bubblewrap-isolated environments. The independent evaluator, which cannot be accessed or modified by the proposer $P$ or $H$, fixes the optimization LLM, EDA flow, synthesis constraints, equivalence checking, scoring objective, and per-design budgets. This isolation ensures a consistent and reliable evaluation protocol while allowing the executable search procedure to evolve. Consequently, neither $P$ nor $H$ can relax verification constraints or alter scoring logic, so all gains can be attributed to improved search.

\subsection{Evidence-Guided Harness Evolution}
\label{sec:method_evolution}

The proposer $P$ is an LLM agent that can read the archive
$\mathcal{A}$, edit harness code, and launch probe runs. At each
round, it selects as parent $H_{\mathrm{par}}$ the harness it
judges most promising, which need not be the highest-scoring
one, based on the implementations, scores, and run records of
all archived harnesses. This lets evolution continue from a
harness whose new mechanism looks promising in its traces
before it pays off in score. Its instructions then prescribe
four stages, shown at the bottom of Fig.~\ref{fig:meta_rtl_overview},
for developing revisions from $H_{\mathrm{par}}$ and
its run records $T_{\mathrm{par}}$.

\textbf{EDA Failure Attribution.}
The proposer inspects $T_{\mathrm{par}}$, including RTL edits,
SEC diagnostics, WNS/TNS/area trajectories, and resource usage,
to trace how candidate transformations fare in
correctness checking and EDA evaluation. It identifies
recurring failures and unproductive search behavior, and
attributes them to the responsible harness decisions, such as
how candidates are constructed or how search state is managed.
A score alone cannot provide this information: a harness whose
edits fail SEC and one whose valid edits bring no gain may both
score zero. The resulting diagnosis names a bottleneck together
with the traces that support it.

\textbf{Harness Hypothesis Generation.}
The proposer turns the diagnosis into a hypothesis: a targeted
change to relevant code paths, prompts, or tool interfaces,
paired with a testable prediction of how search behavior should
change. Typical predictions include more candidates passing SEC,
broader exploration, or fewer unproductive evaluations. Stating
the prediction in advance makes a probe informative even when
the score barely moves, and keeps a lucky gain from being
credited to the change. The proposer then implements it,
and the prediction decides which designs to probe and how to
read the results.

\textbf{Cost-Aware Probing.}
The proposer runs the revised harness on development designs
selected for their relevance to the bottleneck, their ability
to test the prediction, and their evaluation cost. Because
$H_{\mathrm{par}}$ has already run on the same designs, comparing
the new traces with $T_{\mathrm{par}}$ isolates the effect of the
change and shows whether the predicted behavior occurred, for
example in candidate validity, optimization progress, or
resource usage. Each candidate is limited to $K$ probe runs.
Since a full-set evaluation runs every development design under
budget $B$, these probes give an early signal at a
fraction of its cost.

\textbf{Evidence-Based Revision.}
The proposer uses the probe results to accept, refine, or
replace the hypothesis. Confirmed improvements are developed
further, while failed predictions and regressions lead to a
revised implementation or a new hypothesis. Failed predictions
are also informative, since they rule out a suspected cause and
sharpen the next diagnosis. The proposer may cycle through
hypothesis generation, probing, and revision within the $K$-run
limit, but each candidate must end in a submitted harness.

Each round produces up to $C$ candidate harnesses. Every
submitted harness is frozen and formally evaluated on the full
development set $\mathcal{D}_{\mathrm{dev}}$ under budget $B$ to
obtain $Q(H)$, and only harnesses that complete this evaluation
are eligible for selection. The archive records every harness
together with its lineage, score, and run records, so later
rounds build on all accumulated evidence. After $R$ rounds, the
eligible harness with the highest $Q(H)$ is selected as
Meta-RTL$^\ast$. 

\subsection{EDA-Grounded Harness Evaluation}
\label{sec:method_evaluation}

Harness evaluation measures the RTL quality that
a complete optimization run achieves under fixed
resource and EDA constraints. Each run starts from the original
design $D$. The returned RTL $\tilde{D}$ is eligible
for scoring only if it passes correctness checking and yields valid synthesis measurements.
The original design remains available as a fallback
with zero normalized utility.

\textbf{Normalized Timing--Area Utility.}
Meta-RTL uses an equivalent formulation of the
timing--area score in Dr.~RTL~\cite{dr_rtl}.
For a completed run $\tau(H,D,B)$, the normalized
gain in each metric $m$ is defined as
\begin{equation}
    g_m(\tau)
    =
    \frac{m(D)-m(\tilde{D})}{m(D)},
    \quad m\in\{\mathrm{WNS},\mathrm{TNS},A\},
\end{equation}
where WNS, TNS, and $A$ denote worst negative slack,
total negative slack, and area, respectively. Run utility is defined as
\begin{equation}
\begin{aligned}
    q(\tau)
    ={}&
    \alpha g_{\mathrm{WNS}}(\tau)
    +\beta g_{\mathrm{TNS}}(\tau)
    +\gamma g_A(\tau) \\
    &-\lambda\mathbf{1}
    [A(\tilde{D})>(1+\rho)A(D)].
\end{aligned}
\end{equation}
$\alpha$, $\beta$, and $\gamma$ weight the
normalized gains, while $\lambda$ penalizes relative
area growth exceeding $\rho$. Larger utility indicates
greater timing--area optimization gains relative to the original design.

Harness quality is the mean development-set utility:
\begin{equation}
    Q(H)
    =
    \frac{1}{|\mathcal{D}_{\mathrm{dev}}|}
    \sum_{D\in\mathcal{D}_{\mathrm{dev}}}
    q\!\left(\tau(H,D,B)\right).
\end{equation}
Normalization measures improvement relative to
each original design, while averaging summarizes
performance across the development set.
All methods use the same scoring parameters
(Section~\ref{sec:exp_setup}).

\textbf{Resource-Constrained Execution.}
Each inner optimization run receives a budget
$B=(B_{\mathrm{EDA}},B_{\mathrm{LLM}})$, enforcing
\begin{equation}
n_{\mathrm{EDA}} \leq B_{\mathrm{EDA}},
\qquad
c_{\mathrm{LLM}} \leq B_{\mathrm{LLM}},
\end{equation}
where $n_{\mathrm{EDA}}$ counts candidate EDA evaluations
and $c_{\mathrm{LLM}}$ denotes cumulative LLM API cost,
accounting for cached inputs, uncached inputs, and
output tokens.
The shared evaluator enforces both limits independently
of harness decisions.

\begin{table*}[!t]
\centering
\caption{Timing and area comparison on the 20 Dr.~RTL
development designs. Bold indicates the best
unrounded result.}
\label{tab:main_results}
\begingroup
\definecolor{metartlHeader}{HTML}{E4EFF5}
\footnotesize
\setlength{\tabcolsep}{1.5pt}
\renewcommand{\arraystretch}{1.0}
\setlength{\aboverulesep}{0.25ex}
\setlength{\belowrulesep}{0.25ex}

\begin{tabular*}{\textwidth}{
@{\extracolsep{\fill}}
lrrrr@{\hspace{4pt}}
rrrr@{\hspace{4pt}}
rrrr@{}}
\toprule
& \multicolumn{4}{c}{\textbf{Dr.~RTL}}
& \multicolumn{4}{c}{\textbf{RTLScout}}
& \multicolumn{4}{c}{\cellcolor{metartlHeader}\textbf{Meta-RTL}$^\ast$} \\
\cmidrule(lr){2-5}
\cmidrule(lr){6-9}
\cmidrule(l){10-13}
\textbf{Design}
& WNS (ns) & TNS (ns) & $A$ ($\mu$m$^2$) & \multicolumn{1}{c}{$q$}
& WNS (ns) & TNS (ns) & $A$ ($\mu$m$^2$) & \multicolumn{1}{c}{$q$}
& WNS (ns) & TNS (ns) & $A$ ($\mu$m$^2$) & \multicolumn{1}{c}{$q$} \\
\midrule
DSP
& -2.59 & -141.81 & 4127.8 & 0.0527
& -2.57 & -139.78 & 4162.4 & 0.0600
& \textbf{-2.56} & \textbf{-126.17} & \textbf{3996.7} & \textbf{0.0979} \\
FIFO
& -0.40 & -1628.64 & 35613.1 & 0.1968
& -0.41 & -1626.13 & \textbf{34519.1} & 0.1925
& \textbf{-0.26} & \textbf{-942.06} & 37448.5 & \textbf{0.4388} \\
LSTM
& -3.09 & -81.84 & \textbf{3641.3} & 0.5541
& -3.19 & -89.39 & 3863.9 & 0.5283
& \textbf{-2.59} & \textbf{-70.72} & 4972.3 & \textbf{0.6023} \\
SPI
& -0.24 & -24.88 & 1916.0 & 0.2184
& -0.32 & -25.49 & \textbf{1490.4} & 0.1224
& \textbf{-0.22} & \textbf{-22.20} & 1849.5 & \textbf{0.2819} \\
UART
& \textbf{-0.35} & -22.13 & 1134.0 & 0.0714
& -0.37 & -22.48 & 1147.0 & 0.0383
& \textbf{-0.35} & \textbf{-21.36} & \textbf{1001.2} & \textbf{0.0987} \\
aes
& -0.61 & -789.44 & 34532.7 & 0.1084
& -0.70 & -913.49 & 33755.1 & 0.0000
& \textbf{-0.52} & \textbf{-684.54} & \textbf{31071.2} & \textbf{0.2282} \\
arm\_cpu1
& -5.18 & \textbf{-2174.25} & 22520.4 & \textbf{0.1117}
& -5.18 & -2846.87 & 22577.8 & 0.0365
& \textbf{-5.17} & -2224.34 & \textbf{22392.1} & 0.1079 \\
arm\_cpu2
& \textbf{-0.84} & \textbf{-562.75} & 11495.2 & \textbf{0.1256}
& -0.91 & -603.82 & \textbf{11098.3} & 0.0748
& -0.86 & -572.92 & 11715.7 & 0.1072 \\
communication
& -0.36 & -70.71 & 1901.4 & 0.0750
& -0.38 & -71.95 & \textbf{1870.2} & 0.0463
& \textbf{-0.20} & \textbf{-44.50} & 2005.1 & \textbf{0.3931} \\
controller
& -0.38 & -2.85 & 234.6 & 0.0000
& -0.31 & -2.46 & \textbf{231.7} & 0.1419
& \textbf{-0.17} & \textbf{-2.34} & 232.2 & \textbf{0.3405} \\
cpu\_fsm
& -0.57 & -425.66 & 32069.0 & 0.1567
& -0.61 & \textbf{-415.03} & 32256.5 & 0.1401
& \textbf{-0.50} & -433.57 & \textbf{32053.3} & \textbf{0.1930} \\
cpu\_pipe
& \textbf{-2.31} & -260.11 & 5478.0 & 0.0393
& -2.37 & -277.42 & 5349.5 & 0.0094
& \textbf{-2.31} & \textbf{-257.07} & \textbf{5206.4} & \textbf{0.0505} \\
datapath
& -0.83 & -474.59 & 12361.6 & 0.0521
& -0.88 & -513.42 & \textbf{12136.8} & 0.0000
& \textbf{-0.81} & \textbf{-472.61} & 12281.8 & \textbf{0.0658} \\
i2c
& \textbf{-0.31} & -24.98 & 1297.0 & 0.0908
& -0.34 & -25.42 & \textbf{1254.5} & 0.0483
& \textbf{-0.31} & \textbf{-24.52} & 1255.8 & \textbf{0.1016} \\
pcie
& -0.57 & -20.25 & 1037.4 & 0.2697
& -0.52 & -20.30 & \textbf{1006.8} & 0.3027
& \textbf{-0.35} & \textbf{-17.06} & 1036.6 & \textbf{0.4558} \\
router
& \textbf{-0.38} & \textbf{-218.76} & 5530.7 & \textbf{0.2260}
& -0.48 & -271.01 & 5598.8 & 0.0667
& \textbf{-0.38} & -221.17 & \textbf{5432.0} & 0.2258 \\
simple\_spi
& -0.30 & -31.88 & 1227.1 & 0.0450
& -0.32 & -32.02 & \textbf{1194.1} & 0.0164
& \textbf{-0.27} & \textbf{-31.06} & 1210.3 & \textbf{0.1025} \\
ticket\_machine
& -0.10 & \textbf{-0.29} & 31.9 & \textbf{0.6395}
& -0.14 & -0.38 & \textbf{29.5} & 0.5318
& \textbf{-0.08} & -0.45 & 47.1 & 0.6088 \\
tv80
& -1.22 & -362.94 & 5999.9 & 0.0522
& -1.25 & -366.76 & \textbf{5980.5} & 0.0377
& \textbf{-1.10} & \textbf{-342.05} & 6595.5 & \textbf{0.1024} \\
vending\_machine
& \textbf{-0.08} & \textbf{-0.16} & 20470.6 & 0.6471
& -0.09 & \textbf{-0.16} & 20514.5 & 0.6282
& -0.09 & -0.24 & \textbf{12896.2} & \textbf{0.6566} \\
\midrule
Mean $Q$
& -- & -- & -- & 0.1866
& -- & -- & -- & 0.1511
& -- & -- & -- & \textbf{0.2630} \\
\bottomrule
\end{tabular*}
\endgroup
\end{table*}

%% file: sections/experiments.tex
\section{Experiments}
\label{sec:experiments}

\subsection{Experimental Setup}
\label{sec:exp_setup}

We compare Meta-RTL$^\ast$ with two strong LLM-based RTL optimization frameworks: Dr.~RTL~\cite{dr_rtl} with its released skill library and RTLScout~\cite{rtlscout}. For RTLScout, we only evaluate its RTL optimization component while keeping the synthesis flow fixed.

\textbf{Evaluation Benchmarks.}
Our experiments use two benchmark suites,
Dr.~RTL~\cite{dr_rtl} and IC-RTL~\cite{evolve}.
The Dr.~RTL benchmark contains 20 designs
covering arithmetic, control, communication,
and sequential logic, which serve as the
development set.
IC-RTL contains 10 more complex designs
spanning image processing, compression,
geometric computation, and hardware acceleration.
These designs are held out from harness evolution
to evaluate generalization.

\textbf{Outer Evolution.}
Evolution begins with a minimal seed: an in-place RTL
editing loop guided by default EDA reports.
GPT-5.6 Sol~\cite{gpt56sol} serves as the harness proposer,
generating up to $C=2$ candidates per round.
We conduct $R=8$ evolution rounds, yielding $16$ submitted
harnesses in total.
Each candidate is limited to $K=5$ probe runs
on proposer-selected designs.
Submitted harnesses are evaluated afresh on all
20 development designs under the fixed inner budget.
The highest-scoring eligible harness is selected
as Meta-RTL$^\ast$.

\textbf{Inner Optimization.}
All methods share the same EDA flow and use
DeepSeek-V4.1-Flash~\cite{deepseek_v41_flash} as the
optimization model.
Synopsys Design Compiler~\cite{design_compiler}
is used for synthesis and timing analysis,
and Cadence JasperGold~\cite{jaspergold}
for SEC.
To assess RTL-level improvements beyond
standard synthesis optimizations, we synthesize
with a commercial tool and the Nangate
45\,nm~\cite{nangate45} standard-cell library.
Following Dr.~RTL~\cite{dr_rtl}, we impose
a stringent 0.1\,ns clock period to drive
aggressive timing optimization. All methods share per-design budgets of
$B_{\mathrm{EDA}}=50$ EDA evaluations and
$B_{\mathrm{LLM}}\approx\mathrm{US}\$6$ in LLM API
cost.
The 50-evaluation cap matches Dr.~RTL's 10 major rounds
of five minor attempts each.

\textbf{Evaluation Metrics.}
We report worst negative slack (WNS),
total negative slack (TNS), synthesized
area ($A$), and normalized timing--area
utility ($Q$).
The utility metric, defined in
Section~\ref{sec:method_evaluation},
uses $(\alpha,\beta,\gamma)=(0.50,0.35,0.15)$,
$\lambda=0.50$, and $\rho=0.10$,
following Dr.~RTL~\cite{dr_rtl}.
Higher utility indicates better
timing--area optimization.

\textbf{Computational Cost.}
Under DeepSeek-V4.1-Flash pricing,
the estimated costs of optimizing all 20 development
designs are US\$22.5 for Meta-RTL$^\ast$,
US\$75 for Dr.~RTL, and US\$12 for RTLScout. The total recorded LLM API expenditure was US\$627.09,
excluding infrastructure costs (e.g., EDA licenses).

The $C\times R=16$ candidate harnesses yield an estimated evolution cost of US\$510 (16$\times$US\$22.5, i.e., Meta-RTL$^\ast$'s cost per candidate, plus US\$150 for GPT-5.6 Sol proposals).
Amortizing this cost over later applications,
Meta-RTL$^\ast$ would become more cost-efficient
than Dr.~RTL after about 200 design
optimizations, even ignoring the effort
behind the hand-engineered baselines.

\subsection{Development-Set Performance}
\label{sec:exp_main}

Table~\ref{tab:main_results} compares Meta-RTL$^\ast$
with Dr.~RTL and RTLScout on the 20 development
designs. Among the three methods, Meta-RTL$^\ast$
achieves the best or tied-best WNS on 18 designs,
the best TNS on 14 designs, and the lowest area
on 8 designs.

The results show a strong advantage in timing.
Compared with RTLScout, Meta-RTL$^\ast$ achieves
better WNS on 19 designs, ties on one, and
improves TNS on 17 designs, while RTLScout
obtains lower area on 12 designs.
Timing gains also coincide with area reductions:
compared with Dr.~RTL, Meta-RTL$^\ast$
improves or maintains both WNS and TNS while
reducing area on 10 designs, including 7
designs with strict improvements in all three
metrics.

Overall, Meta-RTL$^\ast$ achieves the highest
mean normalized utility of 0.2630, compared
with 0.1866 for Dr.~RTL and 0.1511 for
RTLScout, indicating the strongest aggregate
timing--area performance.

\begin{table}[!t]
\centering
\caption{Timing and area comparison on 10 unseen IC-RTL designs}
\label{tab:generalization}
\begingroup
\definecolor{metartlHeader}{HTML}{E4EFF5}
\small
\setlength{\tabcolsep}{1pt}
\renewcommand{\arraystretch}{1.05}
\setlength{\aboverulesep}{0.25ex}
\setlength{\belowrulesep}{0.25ex}
\begin{tabular*}{\columnwidth}{@{\extracolsep{\fill}}lrrrrrrrr@{}}
\toprule
& \multicolumn{4}{c}{\textbf{Dr.~RTL}}
& \multicolumn{4}{c}{\cellcolor{metartlHeader}\textbf{Meta-RTL}$^\ast$} \\
\cmidrule(lr){2-5}\cmidrule(l){6-9}
\textbf{Design} & WNS & TNS & $A$ & \multicolumn{1}{c}{$q$}
& WNS & TNS & $A$ & \multicolumn{1}{c}{$q$} \\
\midrule
LBP & -0.26 & -22.30 & 1171 & 0.2810 & \textbf{-0.20} & \textbf{-20.41} & \textbf{1092} & \textbf{0.3866} \\
CONV & -2.95 & -165.59 & \textbf{4103} & 0.0909 & \textbf{-2.86} & \textbf{-150.58} & 4166 & \textbf{0.1307} \\
HC & \textbf{-0.62} & \textbf{-111.54} & 2996 & \textbf{0.0552} & -0.63 & -112.25 & \textbf{2962} & 0.0475 \\
JAM & -0.79 & -46.67 & \textbf{1599} & 0.1702 & \textbf{-0.40} & \textbf{-34.46} & 1737 & \textbf{0.4462} \\
DT & -0.65 & -35.57 & 1594 & 0.0439 & \textbf{-0.39} & \textbf{-27.32} & \textbf{1354} & \textbf{0.3326} \\
REF & -6.67 & -1789.23 & 35970 & 0.1726 & \textbf{-3.62} & \textbf{-1311.44} & \textbf{26768} & \textbf{0.4842} \\
GEO & \textbf{-2.85} & \textbf{-699.03} & 14976 & \textbf{0.0534} & -2.97 & -727.84 & \textbf{13207} & 0.0387 \\
IOT & -0.80 & -152.88 & 5448 & 0.3401 & \textbf{-0.73} & \textbf{-110.02} & \textbf{5070} & \textbf{0.4410} \\
SECC & -0.99 & -25.05 & \textbf{1835} & 0.1158 & \textbf{-0.70} & \textbf{-22.59} & 1920 & \textbf{0.2704} \\
GEMM & \textbf{-2.83} & -1573.07 & 41400 & 0.0651 & -2.93 & \textbf{-1455.20} & \textbf{38816} & \textbf{0.0819} \\
\midrule
Mean $Q$ & -- & -- & -- & 0.1388
         & -- & -- & -- & \textbf{0.2660} \\
\bottomrule
\end{tabular*}
\endgroup
\end{table}

\subsection{Generalization to Unseen Designs}
\label{sec:exp_generalization}

To evaluate generalization, we compare
Meta-RTL$^\ast$ with Dr.~RTL on 10 unseen
IC-RTL designs. Both methods leverage
optimization experience from the 20-design
Dr.~RTL benchmark: Dr.~RTL through its
distilled skill library, and Meta-RTL through
harness evolution. For a fair comparison, we exclude RTLScout from the transfer evaluation, as it was not developed on the same benchmark.
The skill library and evolved harness are
kept fixed during evaluation on IC-RTL.

Table~\ref{tab:generalization} presents the
timing and area results. Meta-RTL$^\ast$
achieves better WNS on 7 designs, better
TNS on 8, and lower area on 7.
It also achieves higher utility on 8 of the 10 designs.
Notably, on \textit{REF}, Meta-RTL$^\ast$
simultaneously improves WNS, TNS, and area
by 45.7\%, 26.7\%, and 25.6\%, respectively.
Overall, its mean utility reaches 0.2660,
compared with 0.1388 for Dr.~RTL,
representing a 91.6\% relative improvement.
Note that \textit{REF} contributes substantially to
the aggregate timing gains, yet Meta-RTL$^\ast$
still achieves 79.0\% higher mean utility
on the remaining nine designs.
These results indicate that the evolved
harness transfers effectively to unseen designs.

\subsection{Evolution Analysis}
\label{sec:exp_analysis}

\begin{figure}[t]
  \centering
  \setlength{\abovecaptionskip}{3pt}
  \includegraphics[width=0.84\columnwidth]{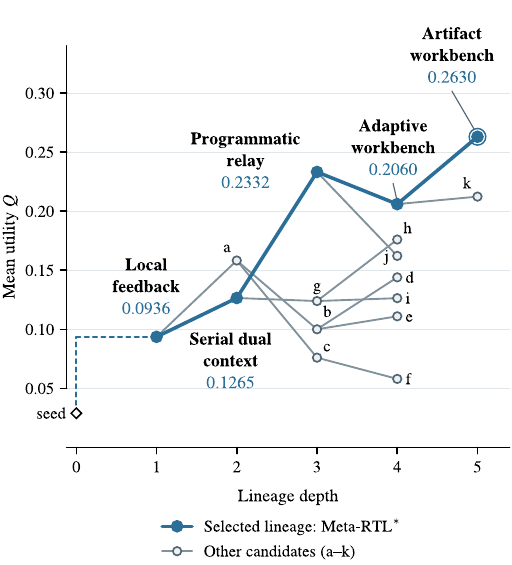}
  \caption{Evolution lineage of Meta-RTL harnesses. }
  \label{fig:evolution_progress}
\end{figure}

\textbf{Evolutionary Progress.}
Fig.~\ref{fig:evolution_progress} traces the
evolution of Meta-RTL$^\ast$.
Starting from the minimal loop,
\textit{Local Feedback} extracts critical-path
information and achieves a utility of 0.0936.
\textit{Serial Dual Context} then improves
utility to 0.1265 by revisiting timing-improved
candidates for area recovery.

The subsequent evolution exemplifies how the proposer
translates execution evidence into harness revisions.
Inspecting \textit{Serial Dual Context} trajectories, the proposer
identified repeated RTL editing failures and premature
search termination that prevented structural transformations
from reaching EDA evaluation. It diagnosed a
delivery\slash continuation bottleneck and hypothesized that
programmatic RTL transformations and bounded search
continuation would enable more valid measurements.
Targeted probing on four development designs increased
mean utility from 0.0817 to 0.2263, with previously
unsuccessful transformations passing SEC. The resulting
\textit{Programmatic Relay} achieved a full-set utility of 0.2332.

Further evolution revealed the overhead of repeatedly
regenerating transformation programs, motivating persistent
artifacts and incremental program edits. Although \textit{Adaptive Workbench} temporarily dipped to 0.2060 due to unpruned workspace state across retries, selecting it as the parent allowed the proposer to isolate this bottleneck and introduce structured artifact indexing, yielding \textit{Artifact Workbench} ($Q=0.2630$, selected as Meta-RTL$^\ast$).

\textbf{Discovered Mechanisms.}
The lineage reveals three
complementary search mechanisms in Meta-RTL$^\ast$.

\emph{EDA-Grounded Transformation Guidance.}
The evolved harness links EDA feedback
to targeted structural transformations.
By interpreting critical-path evidence and
timing--area measurements, it guides the LLM
toward transformations such as arithmetic
sharing, control simplification, and
common-logic factoring.
Subsequent EDA results then refine the
diagnosis, closing an evidence-guided loop.

\emph{Executable Transformation Workspace.}
The evolved harness maintains executable
transformation programs with their
intermediate artifacts and execution feedback
across model requests.
Rather than regenerating RTL edits
from scratch, the optimizer can inspect,
revise, and rerun them.
This enables incremental construction and
debugging of complex structural rewrites,
including repairs of failed attempts.

\emph{Pareto Frontier Memory.}
The evolved harness retains verified RTL
candidates representing different trade-offs
among WNS, TNS, and area.
Rather than committing to a single incumbent,
the optimizer can revisit promising
intermediate candidates and explore
alternative optimization directions.
This preserves opportunities for further
timing or area improvements that might
otherwise be lost during the search.

\subsection{Ablation Studies}
\label{sec:ablations}

\begin{figure}[t]
    \centering
    \setlength{\abovecaptionskip}{3pt}
    \includegraphics[width=0.84\columnwidth]{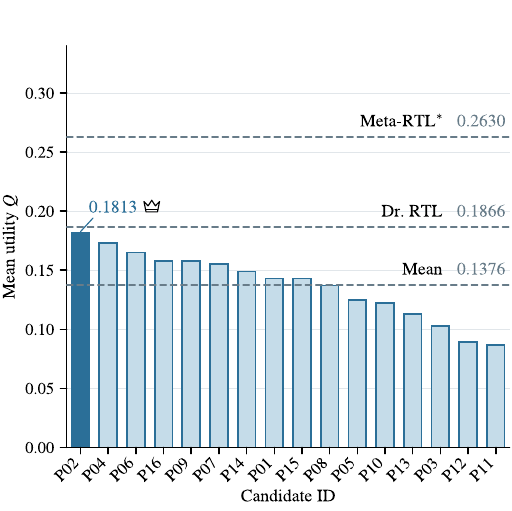}
    \caption{Development-set utility of 16 independently
    designed harnesses.}
    \label{fig:direct_design}
\end{figure}

\textbf{Effect of Harness Evolution.}
To assess the benefit of harness evolution,
we compare Meta-RTL with direct harness design.
Using the same proposer, 16 harnesses are
independently developed from the same minimal seed,
each with up to $K=5$ probe runs but no access
to other candidates or their execution histories.
All are evaluated under the same protocol,
and the best is further tested on unseen designs.

\begin{table}[!t]
\centering
\caption{Ablations on 10 unseen designs (bold: best unrounded)}
\label{tab:ablation}
\begingroup
\definecolor{metartlGreen}{HTML}{E4EFF5}
\scriptsize
\setlength{\tabcolsep}{1.5pt}
\renewcommand{\arraystretch}{1.15}
\resizebox{\columnwidth}{!}{%
\begin{tabular}{@{}l*{11}{c}@{}}
\toprule
\textbf{Config.}
& \textbf{LBP} & \textbf{CONV} & \textbf{HC}
& \textbf{JAM} & \textbf{DT} & \textbf{REF}
& \textbf{GEO} & \textbf{IOT} & \textbf{SECC}
& \textbf{GEMM} & \textbf{Mean} \\
\midrule
\cellcolor{metartlGreen}
Full
& \textbf{0.387} & 0.131 & 0.048
& \textbf{0.446} & \textbf{0.333}
& \textbf{0.484} & 0.039
& \textbf{0.441} & \textbf{0.270}
& \textbf{0.082} & \textbf{0.266} \\
$-$Guidance
& 0.299 & \textbf{0.141} & 0.043
& 0.436 & 0.243 & 0.238
& \textbf{0.056} & 0.407
& 0.153 & 0.047 & 0.206 \\
$-$Frontier
& 0.322 & 0.095 & \textbf{0.048}
& 0.413 & 0.263 & 0.130
& 0.014 & 0.418
& 0.150 & 0.035 & 0.189 \\
$-$Workspace
& 0.362 & 0.027 & 0.038
& 0.308 & 0.316 & 0.122
& 0.014 & 0.385
& 0.124 & 0.012 & 0.171 \\
\bottomrule
\end{tabular}%
}
\endgroup
\end{table}

Fig.~\ref{fig:direct_design} shows the utility
of the 16 independently designed harnesses.
The best achieves 0.1813, below both Dr.~RTL
(0.1866) and Meta-RTL$^\ast$ (0.2630).
Their mean utility is 0.1376.
On unseen designs, the selected harness achieves
0.1050, compared with 0.2660 for Meta-RTL$^\ast$.
These results highlight the benefit of accumulating and organizing
execution evidence across harness revisions.

\textbf{Contributions of Evolved Mechanisms.}
To assess the contributions of the three
discovered mechanisms, we construct three
ablated variants of Meta-RTL$^\ast$ by
individually disabling EDA-grounded
transformation guidance ($-$Guidance),
Pareto frontier memory ($-$Frontier),
and the executable transformation workspace
($-$Workspace).
All variants are evaluated on the same
10 unseen IC-RTL designs under identical
LLM, EDA flow, and budgets, with other
components unchanged.

Table~\ref{tab:ablation} reports the normalized
timing--area utility. Removing Guidance,
Frontier, and Workspace reduces the mean
utility from 0.2660 to 0.2064, 0.1889,
and 0.1707, respectively, corresponding to
decreases of 22.4\%, 29.0\%, and 35.8\%.
The full harness outperforms these variants
on 8, 9, and 10 of the 10 designs, respectively.

Workspace removal causes the largest degradation,
with lower utility on every unseen design.
On \textit{REF}, removing Workspace or Frontier reduces
utility from 0.484 to 0.122 or 0.130,
respectively. Removing Workspace also lowers
utility on \textit{CONV} from 0.131 to 0.027.
These results demonstrate the complementary
roles of retaining executable transformations
for iterative refinement and preserving promising
RTL candidates for continued exploration.

\subsection{A Case Study}
\label{sec:case_study}

\begin{figure}[t]
  \centering
  \setlength{\abovecaptionskip}{3pt}
  \includegraphics[width=\columnwidth]{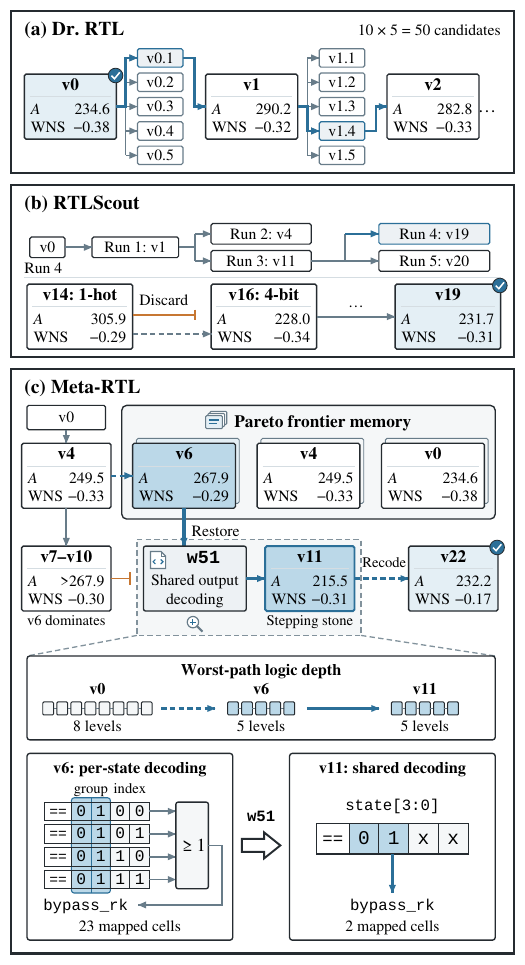}
  \caption{Case study of controller optimization.
TNS is omitted for clarity, as it did not
affect the key trade-offs illustrated here.}
  \label{fig:controller_case}
\end{figure}

Fig.~\ref{fig:controller_case} illustrates, on the
\textit{controller} design, how the discovered mechanisms
interact within the final Meta-RTL$^\ast$ harness.

Dr.~RTL employs repeated five-way candidate expansion and single-candidate promotion, but its search remains focused on expanded decoding without exploring alternative structural directions. Despite evaluating 50 subsequent candidates, it fails to overcome the area penalty, leaving the original v0 as the best design. RTLScout carries the best design from each run into subsequent runs, but abandons the promising one-hot encoding in v14, ultimately settling for a suboptimal four-bit encoding without further refinement.

Through harness evolution, Meta-RTL$^\ast$ gains the ability to identify, preserve, and revisit promising intermediate designs. 
The evolved harness maintains a Pareto frontier
of designs, restores their RTL sources across
optimization contexts, and exposes executable
interfaces for reusing saved transformation programs.
These mechanisms make intermediate designs
reusable starting points for structural refinement,
even when their immediate optimization scores worsen.

Here, v6 uses a two-bit group/two-bit index encoding
to reduce worst-path logic depth from eight to five
levels, but retains per-state output decoding with
a 14.2\% area overhead. The harness preserves v6
in frontier memory, excludes the dominated v7--v10,
and later restores v6 to apply the saved program w51.
By exploiting the common two-bit prefix of four
states asserting \texttt{bypass\_rk}, w51 replaces
their decoding branches with one predicate,
reducing the fan-in cone from 23 to 2 cells and
total combinational cells from 247 to 188.
The resulting v11 passes equivalence checking and
reduces area and WNS/TNS violations by
8.2\%, 18.4\%, and 16.1\%, respectively, relative
to v0, Dr.~RTL's best result. It also matches
RTLScout's best WNS with 7.0\% lower area.
As a structural stepping stone, v11 provides
factored output logic that subsequent candidates
reuse and refine through further state recoding,
ultimately yielding v22 (Table~\ref{tab:main_results}).

%% file: sections/limitations.tex
\section{Limitations and Future Directions}
\label{sec:limitations}

Our results establish the feasibility and
effectiveness of EDA-grounded harness evolution
under a fixed proposer model, optimization model,
EDA flow, and technology library.
However, given the computational budget,
the robustness of these findings to LLM
stochasticity warrants further evaluation
through repeated trials.

Future work could investigate how different
proposer and optimization models shape the
search strategies discovered through evolution.
Evaluating across technology libraries and
physical design flows would further examine
the transferability of these strategies under
different hardware constraints.
Beyond RTL optimization, extending harness
evolution to other EDA tasks may reveal new
mechanisms arising from distinct feedback
signals and search spaces.

%% file: sections/conclusion.tex
\section{Conclusion}
\label{sec:conclusion}

We presented Meta-RTL, which evolves executable
RTL optimization harnesses under a fixed LLM and
EDA flow. Compared with Dr.~RTL, Meta-RTL$^\ast$ reduces
aggregate WNS and TNS violations by 7.8\% and
11.0\% on 20 development designs and transfers
without further evolution to 10 unseen designs,
where it nearly doubles mean utility (0.2660 vs.\ 0.1388).
Analysis and ablations identify EDA-grounded
transformation guidance, an executable transformation
workspace, and a Pareto frontier memory as key mechanisms.
These findings show that EDA feedback can improve
not only RTL designs, but also the search
procedures used to optimize them.